\documentclass{article}
\usepackage{ijcai22}

\usepackage{times}
\usepackage{soul}
\usepackage{url}
\usepackage[hidelinks]{hyperref}
\usepackage[utf8]{inputenc}
\usepackage[small]{caption}
\usepackage{graphicx}
\usepackage{amsmath}
\usepackage{amsthm}
\usepackage{booktabs}
\usepackage{algorithm}
\usepackage{algorithmic}
\graphicspath{{Figures/}}
\usepackage[table]{xcolor}
\usepackage{multirow}
\usepackage[nolist]{acronym}
\usepackage{amssymb}
\usepackage{bm}
\usepackage{makecell}
\usepackage{pifont}% http://ctan.org/pkg/pifont
\newcommand{\cmark}{\ding{51}}

\title{Learning Multi-dimensional Edge Feature-based AU Relation Graph for Facial Action Unit Recognition}

\author{
Cheng Luo$^{1,2,3}$\thanks{Equal contribution.}
\and
Siyang Song$^4$\footnotemark[1]\and
Weicheng Xie$^{1,2,3}$\thanks{Corresponding author.}\and 
Linlin Shen$^{1,2,3}$\And
Hatice Gunes$^4$
\affiliations
$^1$Computer Vision Institute, Shenzhen University\\
$^2$Shenzhen Institute of Artificial Intelligence and Robotics for Society\\
$^3$Guangdong Key Laboratory of Intelligent Information Processing  \\
$^4$Department of Computer Science and Technology, University of Cambridge
\emails
luocheng2020@email.szu.edu.cn, 
ss2796@cam.ac.uk\\
\{wcxie, llshen\}@szu.edu.cn,
Hatice.Gunes@cl.cam.ac.uk
}

\begin{document}

\maketitle

\begin{abstract}

The activations of Facial Action Units (AUs) mutually influence one another. While the relationship between a pair of AUs can be complex and unique, existing approaches fail to specifically and explicitly represent such cues for each pair of AUs in each facial display. This paper proposes an AU relationship modelling approach that deep learns a unique graph to explicitly describe the relationship between each pair of AUs of the target facial display. Our approach first encodes each AU's activation status and its association with other AUs into a node feature. Then, it learns a pair of multi-dimensional edge features to describe multiple task-specific relationship cues between each pair of AUs. During both node and edge feature learning, our approach also considers the influence of the unique facial display on AUs' relationship by taking the full face representation as an input. Experimental results on BP4D and DISFA datasets show that both node and edge feature learning modules provide large performance improvements for CNN and transformer-based backbones, with our best systems achieving the state-of-the-art AU recognition results. Our approach not only has a strong capability in modelling relationship cues for AU recognition but also can be easily incorporated into various backbones. Our PyTorch code is made available at \url{https://github.com/CVI-SZU/ME-GraphAU}.

\end{abstract}

\section{Introduction}

\begin{figure}[t]
  \centering
  \includegraphics[width=1\columnwidth]{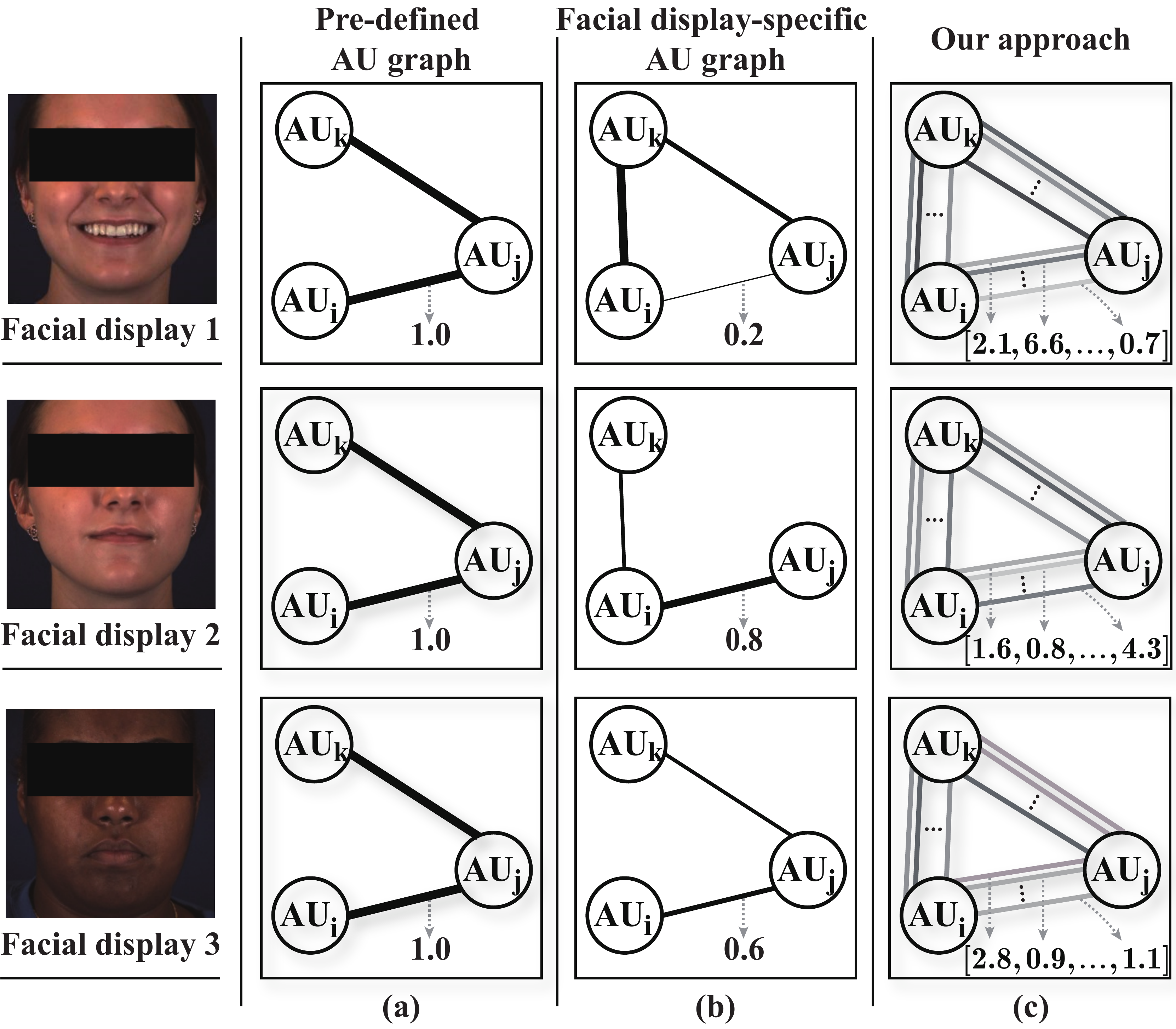}
  \caption{Comparison between our approach with existing AU graph-based approaches: (a) \textbf{pre-defined AU graphs} that use a single topology to define AU association for all facial displays; (b) \textbf{Facial display-specific AU graphs} that assign a unique topology to define AU association for each facial display. Both (a) and (b) use a single value as an edge feature; (c) \textbf{Our approach} encodes a unique AU association pattern for each facial display in node features, and additionally describes the relationship between each pair of AUs using a pair of multi-dimensional edge features.}
\label{fig:intro}
\end{figure}

\noindent Facial Action Coding System~\cite{friesen1978facial} represents human face by a set of facial muscle movements called Action Units (AUs). Compared with the emotion-based categorical facial expression model, AUs describe human facial expressions in a more comprehensive and objective manner~\cite{martinez2017automatic}. Facial AU recognition is a multi-label classification problem as multiple AUs can be activated simultaneously. While previous studies found that underlying relationships among AUs' activation \cite{corneanu2018deep,Song_2021_CVPR,shao2021jaa} are crucial for their recognition, how to properly model such relationships is still an open research question in the field.

A popular strategy applies various traditional machine learning models (\emph{e.g.},  conditional models \cite{eleftheriadis2015multi}) or neural network-based operations (\emph{e.g.}, convolution \cite{zhao2016deep}, Long-Short-Term-Memory networks \cite{niu2019local} or attention \cite{shao2021jaa}), to encode all AU descriptors as a single representation which reflects the underlying relationship among all AUs. A key drawback of such solutions is that they fail to individually model the relationship between each pair of AUs, which may contain crucial cues for their recognition (\textbf{Problem 1}). Some studies represent all AUs of the target face as a graph, where each AU is represented as a node, and each pair of AUs relationship is specifically described by an edge that contains a binary value or a single weight to describe their connectivity or association \cite{song2021uncertain,Song_2021_CVPR}. However, a single value may not be enough to represent the complex underlying relationship between a pair of AUs (\textbf{Problem 2}). In particular, some studies \cite{li2019semantic,liu2020relation} even manually define a single graph topology for all face images based on prior knowledge (\emph{e.g.}, AUs co-occurrence pattern), which fails to consider the influences of the unique facial display on AU relationships (\textbf{Problem 3}).

\begin{figure*}[t]
  \centering
  \includegraphics[width=2.0\columnwidth]{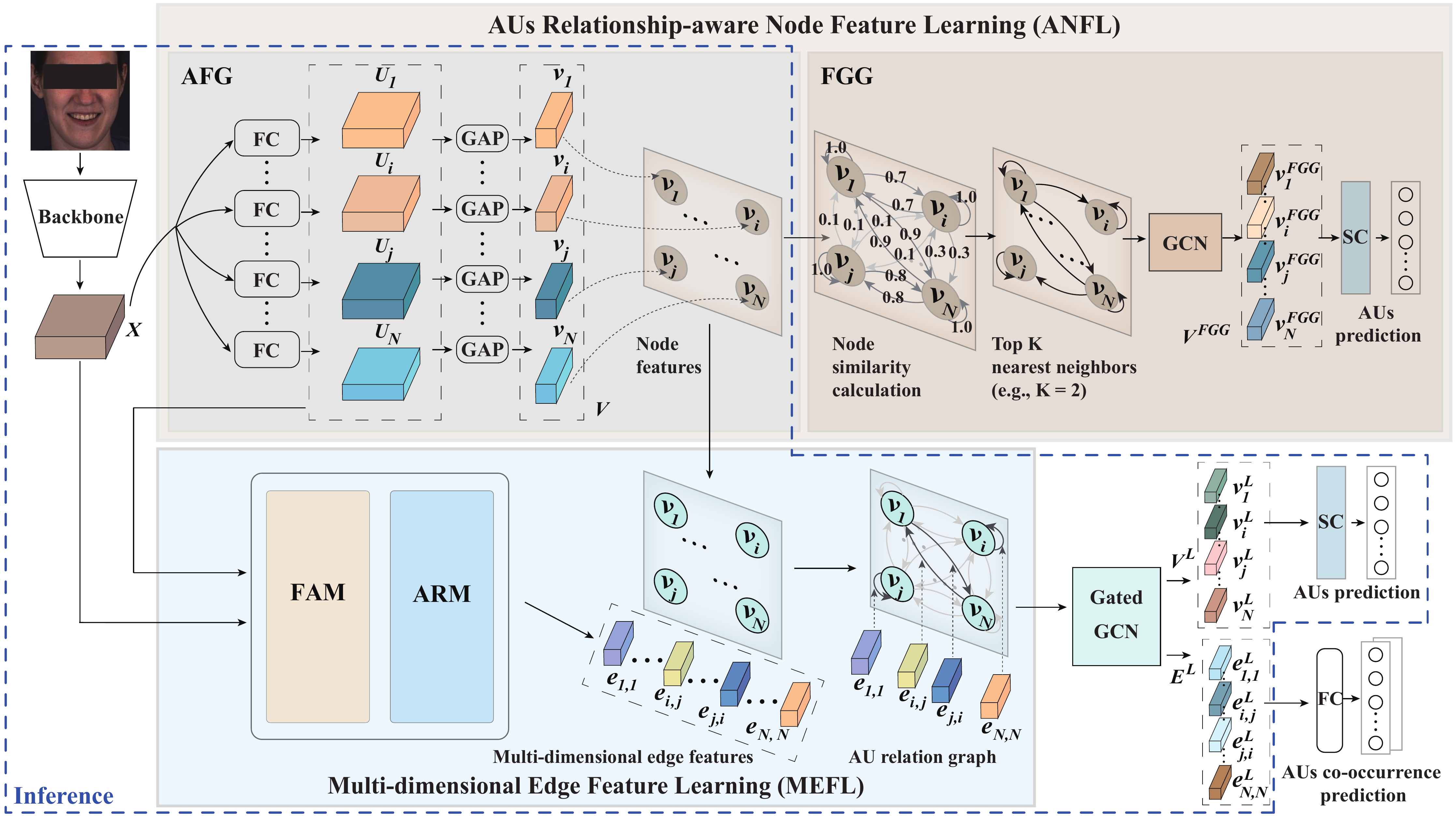}
  \caption{The pipeline of the proposed AU relationship modelling approach. It takes the full face representation $X$ as the input, and the AFG block that is jointly trained with the FGG block, firstly provides a vector as a node feature to describe each AU's activation as well as its association with other AUs (Sec. \ref{subsec: node_feature}). Then, the MEFL module learns a pair of vectors as multi-dimensional edge features to describe task-specific relationship cues between each pair of AUs (Sec. \ref{subsec: multi-dimensional_edge_features}). The AU relation graph produced by our approach is then fed to a GatedGCN for AU recognition. Only the modules and blocks contained within the blue dashed lines are used at the inference stage.}
\label{fig:method_overview}
\end{figure*}

In this paper, we propose a novel AUs relationship modelling approach to address the problems described above, which can be easily incorporated with various deep learning backbones. It takes a full face representation produced by the backbone as the input, and outputs an AUs relation graph that explicitly describes the relationship between each pair of AUs (\textbf{addressing problem 1}). Specifically, our approach consists of two modules: (i) the \textbf{AUs relationship-aware node feature learning (ANFL) module} first individually learns a representation for each AU from the input full face representation (Sec. \ref{subsec: node_feature}), which encodes not only the AU's activation status but also its association with other AUs; and then (ii) the \textbf{multi-dimensional edge feature learning (MEFL) module} learns multiple task-specific relationship cues as the edge representation for each pair of AUs (Sec. \ref{subsec: multi-dimensional_edge_features}) (\textbf{addressing problem 2}). Since both node and edge feature learning take the full face representation as the input, the influence of the unique facial display on AUs relationship is considered when generating its AUs relation graph (\textbf{addressing problem 3}).

In summary, the main contributions of our AU relationship modelling approach are that it represents AU relationships as a unique graph for each facial display, which (i) encodes both the activation status of the AU and its association with other AUs into each node feature; and (ii) learns a multi-dimensional edge feature to explicitly capture the task-specific relationship cues between each pair of AUs. Our multi-dimensional edge encodes unique and multiple relationships between each pair of AUs, rather than a single relationship (\emph{e.g.}, spatial adjacency, co-occurrence patterns, \emph{etc}.) that the single value-edge encoded, which would theoretically generalizes better in modeling complex relationships between vertices \cite{gong2019exploiting,song2021learning,shao2021personality}. The main novelty of the proposed approach in comparison to pre-defined AU graphs \cite{li2019semantic,liu2020relation} and deep learned facial display-specific graphs \cite{song2021uncertain,Song_2021_CVPR} are illustrated in Figure \ref{fig:intro}. To the best of our knowledge, this is the first CNN-GCN approach that conducts end-to-end multi-dimensional edge feature learning for face image processing tasks. The pipeline of the proposed approach is illustrated in Figure \ref{fig:method_overview}. %, and the main contributions of 

\section{The Proposed Approach}

\noindent Our AU relationship modelling approach deep learns a unique AU relation graph from the representation of the target face, which explicitly captures recognition-related relationship cues among AUs based on the end-to-end learned relationship modelling modules. The learned AU relation graph represents the $i_{th}$ AU as the node $\bm{v}_i \in \bm{V}$ in the graph, which contains a vector describing the activation status of the $i_{th}$ AU as well as its association with other AUs in the target facial display. Besides, the task-specific relationship cues between nodes (AUs) $\bm{v}_i$ and $\bm{v}_j$ are also explicitly described by two directed edges $\bm{e}_{i,j}, \bm{e}_{j,i} \in \bm{E}$ that are represented by two deep learned vectors.

\subsection{AUs Relationship-aware Node Feature Learning}
\label{subsec: node_feature}

\noindent As illustrated in Figure~\ref{fig:method_overview}, the ANFL module consists of two blocks: an AU-specific Feature Generator (AFG) and a Facial Graph Generator (FGG). The AFG individually generates a representation for each AU, based on which the FGG automatically designs an optimal graph for each facial display, aiming to accurately recognize all target AUs. To achieve this, the FGG would enforce the AFG to encode task-specific associations among AUs into their AU-specific representations.

\subsubsection{AU-specific Feature Generator} 

The AFG is made up of $N$ AU-specific feature extractors, each of which contains a fully connected layer (FC) and a global average pooling (GAP) layer. It takes the full face representation $\bm{X} \in \mathbb{R}^{D \times C}$ ($C$ channels with $D$ dimensions) as the input, which can be produced by any standard machine learning backbone. The FC layer of $i_{th}$ AU-specific feature extractor first projects the $\bm{X}$ to an AU-specific feature map $\bm{U}_i \in \mathbb{R}^{D \times C}$, which is then fed to a GAP layer, yielding a vector containing $C$ values as the $i_{th}$ AU's representation $\bm{v}_i$. Consequently, $N$ AU representations can be learned from the full face representation $\bm{X}$, respectively.

\subsubsection{Facial Graph Generator}  
\label{subsec: FGG}

\noindent Our hypothesis is that the relationship cues among AUs are unique for each facial display. As a result, directly utilizing relationship cues defined in the training set (\emph{e.g.}, co-occurrence pattern) may not generalise well at the inference stage. As a result, we propose to represent AU relationships in each facial display as a unique graph which considers the influence of the target facial display on AUs relationship.

For a face image, the FGG block treats $N$ target AUs' feature vectors $\bm{\mathcal{V}} = \{\bm{v}_1, \bm{v}_2, \cdots, \bm{v}_N\}$ as $N$ node features and defines the connectivity (edge presence) between a pair of nodes $\bm{v}_i$ and $\bm{v}_j$ by their features' similarity ($\text{Sim}(i,j) = \bm{v}_i^{T}\bm{v}_j$). Specifically, we choose the $K$ nearest neighbours of each node as its neighbours, and thus the graph topology is defined by the learned node features. Then, a GCN layer is employed to jointly update all AUs activation status from the produced graph, where the $i_{th}$ AU's activation representation $\bm{v}_i^{\text{FGG}}$ is generated by $\bm{v}_i$ and its connected nodes as:
 
\begin{equation}
\label{eq:oper_k}
\bm{v}_i^{\text{FGG}} = \sigma[\bm{v}_i + g(\bm{v}_i, \sum_{j=1}^N r(\bm{v}_j, a_{i,j}))], 
\end{equation}
where $\sigma[]$ is the non-linear activation; $g$ and $r$ denote differentiable functions of the GCN layer, and $a_{i,j} \in \{0, 1\}$ represents the connectivity between $\bm{v}_i$ and $\bm{v}_j$. 

To provide a prediction for the $i_{th}$ AU, we propose a similarity calculating (SC) strategy which learns a trainable vector $\bm{s}_i$ that has the same dimension as the $\bm{v}_i^{\text{FGG}}$, and then generates the $i_{th}$ AU's occurrence probability by computing the cosine similarity between $\bm{v}_i^{\text{FGG}}$ and $\bm{s}_i$ as:
\begin{equation}
\label{eq:SC}
    p_i^{\text{FGG}} = \frac{\text{ReLU}(\bm{v}_i^{\text{FGG}})^T\text{ReLU}(\bm{s}_i)}{\| \text{ReLU}(\bm{v}_i^{\text{FGG}}) \|_2  \|\text{ReLU}(\bm{s}_i)\|_2},
\end{equation}
where ReLU denotes a non-linearity activation. As a result, a pair of AUs that have a strong association (high similarity) would have connected nodes. In other words, the FGG block enforces the AFG block to encode node (AU) features that contain task-specific relationship cues among AUs of the target facial display, in order to produce an optimal graph for their recognition.

\subsection{Multi-dimensional Edge Feature Learning}
\label{subsec: multi-dimensional_edge_features}

\begin{figure}[htb]
  \centering
  \includegraphics[width=1\columnwidth]{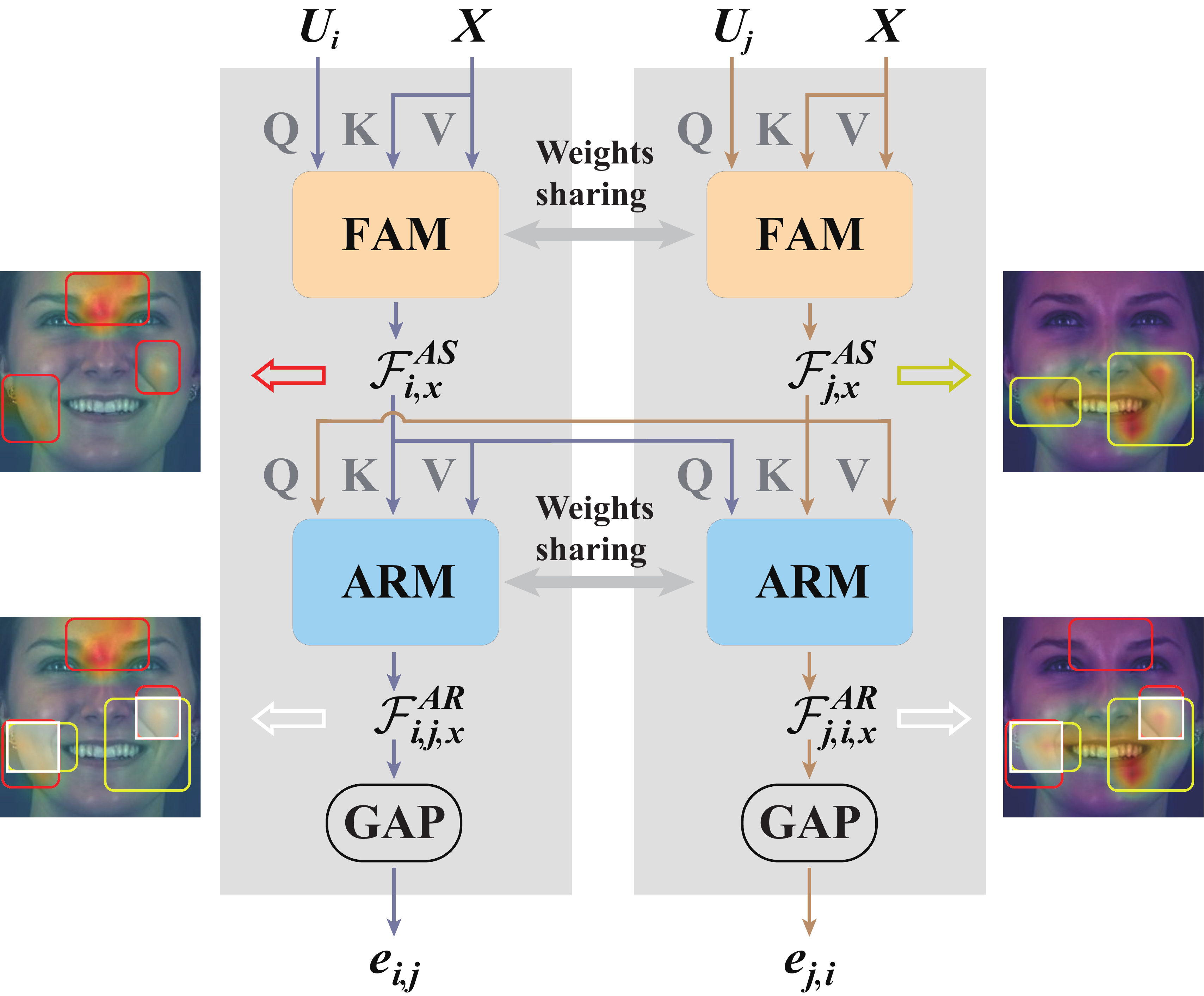}
  \caption{Illustration of the MEFL module. The \textbf{FAM} first independently locates activation cues related to $i_{th}$ and $j_{th}$ AU-specific feature maps $\bm{U}_i$ and $\bm{U}_j$ in the full face representation $\bm{X}$ (activated face areas are depicted in red and yellow). Then, the \textbf{ARM} further extracts cues related to both $\bm{U}_i$ and $\bm{U}_j$ (depicted in white), based on which multi-dimensional edge features $\bm{e}_{i,j}$ and $\bm{e}_{j,i}$ are produced.}
\label{fig:method_GEM}
\end{figure}

\noindent In addition to relationship cues encoded in node features, we also propose a Multi-dimensional Edge Feature Learning (MEFL) module to deep learn a pair of multi-dimensional edge features, aiming to explicitly describe task-specific relationship cues between each pair of AUs. Importantly, we learn edge features for both connected and un-connected node pairs defined in Sec. \ref{subsec: node_feature}. Even when a pair of nodes have low similarity, their relationship may still contain crucial cues for AU recognition, which are ignored during the node feature learning. Since an AU's activation may also influence other AUs' status, the relationship between a pair of AUs can be reflected by not only their features but also AUs defined by other facial regions. Thus, the MEFL module consists of two blocks: a \textbf{Facial display-specific AU representation modelling (FAM)} block that first locates activation cues of each AU from the full face representation, and an \textbf{AU relationship modelling (ARM)} block that further extracts features from these located cues, which relate to both AUs activation. This is also illustrated in Figure \ref{fig:method_GEM}.

\paragraph{FAM.} As illustrated in Figure \ref{fig:method_GEM}, for a pair of AUs, the FAM takes their AU-specific feature maps $\bm{U}_i$, $\bm{U}_j$, and the full face representation $\bm{X}$ as the input. It first conducts cross attention between $\bm{U}_i$ and $\bm{X}$ as well as $\bm{U}_j$ and $\bm{X}$, respectively, where AU-specific feature maps $\bm{U}_i$ and $\bm{U}_j$ are individually used as queries, while the full face representation $\bm{X}$ is treated as the key and value. This process can be formulated as:
\begin{equation}
    \bm{\mathcal{F}}^{AS}_{i,x}, \bm{\mathcal{F}}^{AS}_{j,x} = \text{FAM}(\bm{U}_i,\bm{X}), \text{FAM}(\bm{U}_j,\bm{X}), \\     
\end{equation}
with the cross attention operation in FAM defined as:
\begin{equation}
     \text{FAM}(\bm{A}, \bm{B}) = \text{softmax}(\frac{\bm{A} \bm{W}_q (\bm{B} \bm{W}_k)^T}{\sqrt{d_k} })\bm{B} \bm{W}_v,   
\label{eq:fam}
\end{equation}
where $\bm{W}_q$, $\bm{W}_k$ and $\bm{W}_v$ are learnable weights, and $d_k$ is a scaling factor equalling to the number of the 'key's' channels. As a result, the produced $\bm{\mathcal{F}}^{AS}_{i,x}$ and $\bm{\mathcal{F}}^{AS}_{j,x}$ extract and highlight the most important facial cues from all facial regions of the target facial display for AU $i$ and AU $j$'s recognition, respectively, which consider the influence of the unique facial display on AUs relationships.

\paragraph{ARM.} After encoding task-specific facial cues for each AU's recognition independently, the ARM block further extracts the facial cues related to both AUs' recognition. It also conducts the cross-attention (has the same form as Eq.~\ref{eq:fam} but independent weights) between $\bm{\mathcal{F}}^{AS}_{i,x}$ and $\bm{\mathcal{F}}^{AS}_{j,x}$, and produces features $\bm{\mathcal{F}}^{AR}_{i,j,x}$ and $\bm{\mathcal{F}}^{AR}_{j,i,x}$, where $\bm{\mathcal{F}}^{AR}_{i,j,x}$ is generated by using $\bm{\mathcal{F}}^{AS}_{j,x}$ as the query and $\bm{\mathcal{F}}^{AS}_{i,x}$ as the key and value, while $\bm{\mathcal{F}}^{AR}_{j,i,x}$ is generated by using $\bm{\mathcal{F}}^{AS}_{i,x}$ as the query and $\bm{\mathcal{F}}^{AS}_{j,x}$ as the key and value. As a result, the $\bm{\mathcal{F}}^{AR}_{i,j,x}$ summarizes $\bm{\mathcal{F}}^{AS}_{j,x}$-related cues in the $\bm{\mathcal{F}}^{AS}_{i,x}$, and $\bm{\mathcal{F}}^{AR}_{j,i,x}$ summarizes $\bm{\mathcal{F}}^{AS}_{i,x}$-related cues in the $\bm{\mathcal{F}}^{AS}_{j,x}$. Finally, we feed $\bm{\mathcal{F}}^{AR}_{i,j,x}$ and $\bm{\mathcal{F}}^{AR}_{j,i,x}$ to a GAP layer to obtain multi-dimensional edge feature vectors $\bm{e}_{i,j}$ and $\bm{e}_{j,i}$, respectively. Mathematically speaking, this process can be represented as
\begin{equation}
    \bm{e}_{i,j}, \bm{e}_{j,i} = \text{GAP}(
    \text{ARM}(\bm{\mathcal{F}}^{AS}_{j,x},\bm{\mathcal{F}}^{AS}_{i,x}),
    \text{ARM}(\bm{\mathcal{F}}^{AS}_{i,x},\bm{\mathcal{F}}^{AS}_{j,x})
    ).     
\end{equation}
In short, the features encoded in edge features $\bm{e}_{i,j}$ and $\bm{e}_{j,i}$ summarize multiple facial cues that relate to both $i_{th}$ and $j_{th}$ AUs' recognition, from all facial regions of the target face.

Once the AUs relation graph $\bm{G}^0=(\bm{V}^0,\bm{E}^0)$ that consists of $N$ node features and $N \times N$ multi-dimensional directed edge features is learned, we feed it to a GCN model to jointly recognize all target AUs. In this paper, we use a model that only consists of $L$ gated graph convolution layers (GatedGCN) \cite{bresson2017residual}, and thus the output $\bm{G}^L=(\bm{V}^L,\bm{E}^L)$ is also a graph that has the same topology as $\bm{G}^0$, where the $i_{th}$ node $\bm{v}_i^L$ represents the $i_{th}$ AU's activation status ($L = 2$ in this paper). We finally re-employ the SC module proposed in the FGG block to predict $N$ AUs' activation from the node features of $\bm{G}^L$. During the inference stage, only the well-trained AFG and MEFL are used to process the input full face representation and generate the AU relation graph.

\begin{table*}[thb]

    \centering

     \small
     \setlength{\tabcolsep}{1.25mm}{
        \begin{tabular}{lccccccccccccc}

        \Xhline{3\arrayrulewidth}
        \multicolumn{1}{l}{\multirow{2}{*}{Method}} & \multicolumn{12}{c}{AU}  & \multirow{2}{*}{\textbf{Avg.}} \\ \cmidrule(lr){2-13}
        \multicolumn{1}{l}{}    & 1    & 2    & 4    & 6   & 7   & 10   & 12   & 14   & 15   & 17   & 23   & 24   &      \\ \midrule
        DRML  \cite{zhao2016deep}                  &36.4  &41.8  &43.0  &55.0   &67.0   &66.3  &65.8   &54.1   &33.2  &48.0  &31.7  &30.0  &48.3 \\
        EAC-Net \cite{li2018eac}                &39.0  &35.2  &48.6  &76.1   &72.9   &81.9  &86.2   &58.8   &37.5  &59.1  &35.9  &35.8  &55.9  \\
        JAA-Net  \cite{shao2018deep}               &47.2  &44.0  &54.9  &77.5   &74.6   &84.0  &86.9   &61.9   &43.6  &60.3  &42.7  &41.9  &60.0  \\

        LP-Net  \cite{niu2019local}                &43.4  &38.0  &54.2  &77.1   &76.7   &83.8  &87.2   &63.3   &45.3  &60.5  &48.1  &54.2  &61.0  \\
        ARL   \cite{shao2019facial}                  &45.8  &39.8  &55.1  &75.7   &77.2   &82.3  &86.6   &58.8   &47.6  &62.1  &47.4  &[55.4]  &61.1  \\
        SEV-Net \cite{yang2021exploiting}                &[\textbf{58.2}]  &[\textbf{50.4}]  &58.3  &[\textbf{81.9}]  &73.9   &[\textbf{87.8}]  &87.5   &61.6   &[52.6]  &62.2  &44.6  &47.6  &63.9  \\
        FAUDT \cite{jacob2021facial}                   &51.7  &[49.3]  &[\textbf{61.0}]  &77.8   &\underline{79.5}   &82.9  &86.3   &[67.6]   &51.9  &63.0  &43.7  &[\textbf{56.3}]  &\underline{64.2}   \\\midrule
    
        SRERL \cite{li2019semantic}                  &46.9  &45.3  &55.6  &77.1   &78.4   &83.5  &\underline{87.6}   &63.9   &52.2  &[63.9]  &47.1  &53.3  &62.9  \\
        UGN-B \cite{song2021uncertain}                  &[54.2]  &46.4  &56.8  &76.2   &76.7   &82.4  &86.1   &64.7   &51.2  &63.1  &48.5  &53.6  &63.3  \\
        HMP-PS \cite{Song_2021_CVPR}                 &53.1  &46.1  &56.0  &76.5   &76.9   &82.1  &86.4   &64.8   &51.5  &63.0  &[49.9]  &54.5  &63.4  \\ \midrule
        
        \rowcolor{gray!30}

        Ours (ResNet-50)           &\underline{53.7}      &\underline{46.9}      &\underline{59.0}      &\underline{78.5}       &[80.0]       &\underline{84.4}      &[87.8]       &\underline{67.3}       &\underline{52.5}   &\underline{63.2}   &\textbf{50.6}  &52.4  &[64.7]     \\\rowcolor{gray!30}
        Ours (Swin Transformer-Base)              &52.7  &44.3  &[60.9]  & [79.9]   &[\textbf{80.1}]   &[85.3]  &[\textbf{89.2}]   &[\textbf{69.4}]   &[\textbf{55.4}]  &[\textbf{64.4}]  &\underline{49.8}  &\underline{55.1}  &[\textbf{65.5}]  \\
        \Xhline{3\arrayrulewidth}
        \end{tabular}}

    \caption{F1 scores (in \%) achieved for 12 AUs on BP4D dataset, where the three methods (SRERL, UGN-B and HMP-PS) listed in the middle of the table are also built with graphs. The best, second best, and third best results of each column are indicated with brackets and bold font, brackets alone, and underline, respectively.
    }
    \label{ex:tab_BP4D_sota}
\end{table*}

\begin{table*}[thb]
    \centering

    \small
    \setlength{\tabcolsep}{1.3mm}{
    \begin{tabular}{lccccccccc}
    \Xhline{3\arrayrulewidth}
    \multicolumn{1}{l}{\multirow{2}{*}{Method}} & \multicolumn{8}{c}{AU}  & \multirow{2}{*}{\textbf{Avg.}} \\ \cmidrule(lr){2-9}
    \multicolumn{1}{l}{}    & 1               & 2       & 4   & 6   & 9   & 12   & 25   & 26     &      \\\midrule
    DRML \cite{zhao2016deep}                    &17.3             &17.7     &37.4  &29.0   &10.7   &37.7  &38.5   &20.1   &26.7 \\
    EAC-Net \cite{li2018eac}                 &41.5             &26.4     &66.4  &50.7   &[\textbf{80.5}]   &[\textbf{89.3}]  &88.9   &15.6   &48.5   \\
    JAA-Net \cite{shao2018deep}                 &43.7             &46.2     &56.0  &41.4   &44.7   &69.6  &88.3   &58.4   &56.0   \\

    LP-Net \cite{niu2019local}                 &29.9              &24.7              &72.7  &46.8   &49.6   &72.9  &\underline{93.8}   &\underline{65.0}   &56.9   \\
    ARL \cite{shao2019facial}                    &43.9              &42.1              &63.6  &41.8   &40.0   &\underline{76.2}  &[95.2]   &[66.8]   &58.7   \\
    SEV-Net  \cite{yang2021exploiting}                &\textbf{[55.3]}     &[\textbf{53.1}]     &61.5  &\underline{53.6}   &38.2   &71.6  &[\textbf{95.7}]   &41.5   &58.8   \\
    FAUDT \cite{jacob2021facial}                  &46.1              &[48.6]  &\underline{72.8}  &[\textbf{56.7}]   &50.0   &72.1  &90.8   &55.4   &\underline{61.5}   \\\midrule
    
    SRERL \cite{li2019semantic}                  &45.7              &47.8  &59.6  &47.1   &45.6   &73.5  &84.3   &43.6   &55.9   \\
    UGN-B \cite{song2021uncertain}                  &43.3              &\underline{48.1}  &63.4  &49.5   &48.2   &72.9  &90.8   &59.0   &60.0   \\
    HMP-PS  \cite{Song_2021_CVPR}                 &38.0              &45.9  &65.2  &50.9   &\underline{50.8}   &76.0  &93.3   &[\textbf{67.6}]   &61.0   \\ \midrule
    
    \rowcolor{gray!30}
    Ours (ResNet-50)              &[54.6]  &47.1  &[72.9]  &[54.0]   &[55.7]   &[76.7]  &91.1   &53.0  &[\textbf{63.1}]     \\\rowcolor{gray!30}
    Ours (Swin Transformer-Base)              &\underline{52.5}     &45.7      &[\textbf{76.1}]      &51.8       &46.5       &76.1      & 92.9      &57.6      &[62.4]    \\
    \Xhline{3\arrayrulewidth}
    \end{tabular}
    }
    \caption{F1 scores (in \%) achieved for 8 AUs on DISFA dataset. }
    \label{ex:tab_DISFA_sota}
\end{table*}

\subsection{Training Strategy}

\noindent In this paper, we propose a two-stage training method to jointly optimize the proposed ANFL and MEFL modules with the backbone and classifier in an end-to-end manner.

In the first stage, we train the backbone with the ANFL module, aiming to learn an AFG block that produces node features containing both AU activation status and their associations for each facial display. A priori, we notice that existing AU datasets usually have imbalanced labels, where some AUs occurred less frequently than others, and most AUs are inactivated for the majority of face images. To alleviate such issues, we propose a weighted asymmetric loss to compute the loss value between the ground-truth and predictions generated by the FGG block. It is inspired by the asymmetric loss \cite{ridnik2021asymmetric}, but has a unique weight for each sub-task (each AU's recognition) as well as fewer hyperparameters. The proposed weighted asymmetric loss is formulated as:

\begin{equation}
\label{eq:AUR_loss}
    \mathcal{L}_{\text{WA}} = -\frac{1}{N}\sum_{i = 1}^{N} w_i [y_i \text{log}(p_i)+(1-y_i) p_i \text{log}(1-p_i)],
\end{equation}
where $p_i$, $y_i$ and $w_i$ are the prediction (occurrence probability), ground truth and weight of the $i_{th}$ AU, respectively. Here, the $w_i = N(1/r_i)/\Sigma_{j=1}^N(1/r_j)$ is defined by the $i_{th}$ AU's occurrence rate $r_i$ computed from the training set. It allows loss values to account less for AUs that have higher occurrence rates in the training set, leading loss values caused by less frequently occurring AUs to have higher weights during the training. Additionally, the term '$p_i$' in the center of $(1-y_i) p_i \text{log}(1-p_i)$ down weights loss values caused by inactivated AUs that are easy to be recognized, whose predicted occurrence probabilities are close to zero ($p_i \ll 0.5$), enforcing the training process to focus on activated AUs and inactivated AUs that are hard to be correctly recognized.

The second stage trains the MEFL module and classifier (GatedGCN) with the pre-trained backbone and AFG block. Here, we again employ the proposed weighted asymmetric loss (Eq.~\ref{eq:AUR_loss}) to compute the loss value $\mathcal{L}_{\text{WA}}$ between the outputs of the classifier and ground truth. Additionally, we also leverage the AUs co-occurrence patterns to supervise the training process. We feed multi-dimensional edge features $\bm{e}_{i,j}^L$ and $\bm{e}_{j,i}^L$ generated from the last GatedGCN layer to a shared FC layer, in order to predict the co-occurrence pattern of the $i_{th}$ and $j_{th}$ AUs of the target face. We define this task as a four-class classification problem, \emph{i.e.,} for a pair of nodes $\bm{v}_i$ and $\bm{v}_j$: (1) both $\bm{v}_i$ and $\bm{v}_j$ are inactivated; (2) $\bm{v}_i$ is inactivated and $\bm{v}_j$ is activated; (3) $\bm{v}_i$ is activated and $\bm{v}_j$ is inactivated; or (4) both $\bm{v}_i$ and $\bm{v}_j$ are activated. As a result, the categorical cross-entropy loss is introduced as:
\begin{equation}
    \mathcal{L}_{\text{E}} = -\frac{1}{| \bm{E} |}\sum_{i = 1}^{| \bm{E} |} \sum_{j = 1}^{N_E} y_{i,j}^{e} \text{log}(\frac{e^{p_{i,j}^{e}}}{\sum_{k}e^{p_{i,k}^{e}}}),
\end{equation}
where $| \bm{E} |$ denotes the number of edges in the facial graph; $N_E$ is the number of co-occurrence patterns; $p_{i,j}^{e}$ is the co-occurrence prediction output from the shared FC layer. Consequently, The overall training loss of the second stage is formulated as the weighted combination of the two losses:
\begin{equation}
    \mathcal{L} =  \mathcal{L}_{\text{WA}} + \lambda \mathcal{L}_{\text{E}},
    \label{method:loss_function}
\end{equation}
where $\lambda$ decides the relative importance of the two losses.

\section{Experiments}

\subsection{Experimental Setup}

\paragraph{Datasets.} We evaluate the performance of our approach on two widely-used benchmark datasets: BP4D \cite{zhang2014bp4d} and DISFA \cite{mavadati2013disfa}. BP4D recorded 328 videos (about 140,000 facial frames) from 41 young adults (23 females and 18 males) who were asked to respond to $8$ emotion elicitation tasks. DISFA recorded $130,815$ frames from 27 subjects (12 females and 15 males) who were watching Youtube videos. Each frame in BP4D and DISFA is annotated with occurrence labels of multiple AUs.

\paragraph{Implementation Details.} 

For both datasets, we use MTCNN \cite{yin2017multi} to perform face detection and alignment for each frame and crop it to $224 \times 224$ as the input for backbones. We then follow the same protocol as previous studies \cite{zhao2016deep,li2018eac,Song_2021_CVPR} to conduct subject-independent three folds cross-validation for each dataset, and report the average results over 3 folds. During the training, we employ an AdamW optimizer with $\beta_1 = 0.9$, $\beta_2 = 0.999$ and weight decay of $5e^{-4}$. The number K for choosing nearest neighbors in the FGG is set to 3 and 4 for BP4D and DISFA, respectively. For the hyperparameter $\lambda$ in Eq.~\ref{method:loss_function}, we set it to 0.05 and 0.01 for models based on ResNet and Swin Transformer, respectively. We totally train the proposed model for 40 epochs, including 20 epochs for the first stage (the initial learning rate of $1e^{-4}$) and 20 epochs for the second stage (the initial learning rate of $1e^{-6}$), with a batch size of 64. The cosine decay learning rate scheduler is also used. Both backbones are pre-trained on ImageNet \cite{deng2009imagenet}.

\paragraph{Evaluation Metric.} We follow previous AU occurrence recognition studies \cite{shao2021jaa,churamani2021aula,li2019self,Song_2021_CVPR} using a common metric: frame-based F1 score, to evaluate the performance of our approach, which is denoted as $F1 = 2 \frac{P \cdot R}{P+R}$. It takes the recognition precision $P$ and recall rate $R$ into consideration.

\subsection{Results and Discussion}
\label{subsec:results}

\paragraph{Comparison to State-of-the-art Methods.} This section compares our best systems of two backbones with several state-of-the-art methods on both datasets. Table~\ref{ex:tab_BP4D_sota} reports the occurrence recognition results of 12 AUs on BP4D. We additionally provide the AUC results in Appendix \ref{sec:ex_ar}. It can be observed that the proposed AU relationship modelling approach allows both backbones (ResNet-50 and Swin Transformer-Base (Swin-B)) to achieve superior overall F1 scores than all other listed approaches, with $0.5\%$ and $1.3\%$ average improvements over the state-of-the-art \cite{jacob2021facial}. Specifically, our approach allows both backbones to achieve the top three performances for $9$ out of $12$ AUs' recognition (\emph{e.g.}, AU 4, AU 6, AU 7, AU 10, AU 12, AU 14, AU 15, AU 17, and AU 23) among all listed approaches. Similar results were also achieved on DISFA. According to Table~\ref{ex:tab_DISFA_sota}, our approach helps both backbones to achieve the state-of-the-art average F1 scores over $8$ AUs, which outperform the current state-of-the-art with 1.6\% and 0.9\% improvements, respectively. For fair comparisons, we only compare our approach with static face-based methods that did not remove any frame from the datasets.

\begin{figure}[t]
  \centering
  \includegraphics[width=1.0\columnwidth]{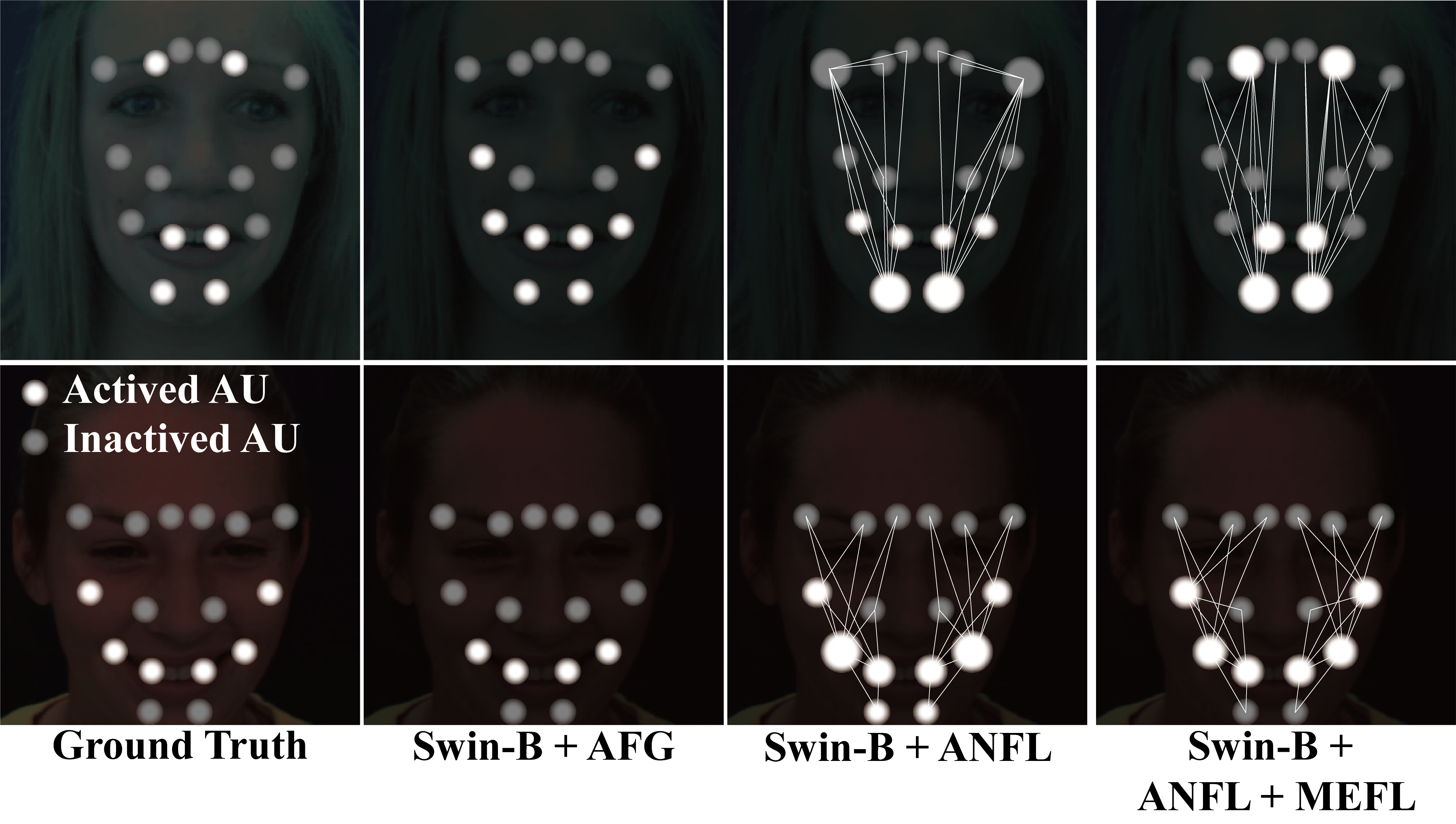}
  \caption{Visualization of association cues encoded in node features (only systems of the last two columns encode such cues). We connect each node to its K nearest neighbours, where nodes of activated AUs usually have more connections than nodes of inactivated AUs. Systems used such relationship cues have enhanced AU recognition results (predictions of the column 3 is better than the column 2).}
\label{fig:ex_visualization}
\end{figure}

According to both tables, our ResNet-50-based system also clearly outperforms other graph-based AU recognition approaches which also use CNNs (ResNet (UGN-B, HMP-PS) or VGG (SRERL)) as backbones. Since SRERL only uses a pre-computed adjacent matrix to describe the relationship between AUs for all faces, our system shows a large advantage over it, with $1.8\%$ and $7.2\%$ F1 score improvements for the average results on BP4D and DISFA, respectively. Although UGN-B and HMP-PS assigned each facial display a unique adjacent matrix and achieved better performance than SRERL, they still use a single value to describe the relationship between each pair of AUs, without considering multiple relationship cues. Thus, our deep-learned task-specific multi-dimensional edge features lead our system to achieve more than $1.3\%$ and $2.1\%$ average F1 score improvements over UGN-B and HMP-PS on both datasets.

\begin{table}[t]
    \centering
    \small
    \setlength{\tabcolsep}{1.8mm}{
    \begin{tabular}{cccc|cc|rr}
    \Xhline{3\arrayrulewidth}
     Backbone  &AFG   &FGG  &MEFL  & $\mathcal{L}_{\text{WA}}$ & $\mathcal{L}_{\text{E}}$ & \textbf{Res} &\textbf{Swin}\\
    \midrule
     \cmark &        &        &        &        &        &59.1 &62.6    \\
     \cmark & \cmark &        &        &        &        &60.4 &63.6     \\
      \cmark &       &        &        & \cmark &        &61.8 &63.9     \\
    \cmark & \cmark &\cmark        &        &    &        &63.1 &63.6     \\
    \cmark & \cmark &         &\cmark        &    &        &63.2  &63.8     \\
    % \cmark & \cmark &\cmark   &\cmark        &    &        &63.9  &-     \\

     \cmark & \cmark &        &        & \cmark &        &63.0 &64.6     \\
     \cmark & \cmark & \cmark &        & \cmark &        &63.7 &65.1   \\
     \cmark & \cmark &        & \cmark & \cmark &        &63.9 &64.6    \\     
     \cmark & \cmark & \cmark & \cmark & \cmark &        &64.5 &65.4    \\
     \cmark & \cmark &\cmark  & \cmark & \cmark & \cmark &64.7 &65.5   \\
   \Xhline{3\arrayrulewidth}
    \end{tabular}
    }
    \caption{Average AU recognition results (F1 scores (in \%)) achieved by various settings using two backbones on the BP4D. The systems of the first two rows are trained with widely-used weighted binary cross-entropy loss.}
    \label{ex:tab_AblationStudy}
\end{table}

\paragraph{Ablation Studies.} Table~\ref{ex:tab_AblationStudy} evaluates the influence of each component of our pipeline on the average AU recognition results. It can be observed that simply using the AFG block to specifically learn a representation for each AU enhanced the performance for both backbones, indicating that the relationship between each AU's activation and the full face representation is unique. In particular, when a facial AU is activated, its movement usually affects other facial regions (\emph{i.e.}, the activation of other AUs) while inactivated AUs would not have such an effect. As visualized in Figure \ref{fig:ex_visualization}, our FGG simulates this phenomenon by connecting activated AUs to all other AUs (including activated and inactivated AUs). Building on the backbone-AFG system, we also found that individually adding the FGG block or MEFL module further increased the recognition performance for both backbones. These results suggest that (i) the FGG block allows the AFG block to encode additional AU recognition-related cues into node features, \emph{i.e.,} we hypothesize that the FGG can help the AFG to learn AUs' relationship cues for their recognition; and (ii) the multi-dimensional edge features learned by the MEFL module provide more task-specific AU relationship cues to improve the recognition performance, which further validates our hypothesis that a single value is not enough to carry all useful relationship cues between a pair of AUs.

In short, the proposed approach can provide valuable relationship cues for AU recognition during both node and edge feature learning. More importantly, jointly using FGG and MEFL with our weighted asymmetric loss largely boosted both backbones' recognition capabilities, \emph{i.e.}, $5.6\%$ and $2.9\%$ F1 score improvements over the original backbones, as well as $1.7\%$ and $0.9\%$ improvements over the backbone-AFG systems. Besides the proposed relationship modelling approaches, we show that the two loss functions also positively improved the recognition performance. The weighted asymmetric loss clearly enhanced the performance over the widely-used weighted binary cross-entropy loss, illustrating its superiority in alleviating data imbalance issue. Meanwhile, the proposed AU co-occurrence supervision also slightly enhanced recognition results for both backbones.

\section{Conclusion}

\noindent This paper proposes to deep learn a graph that explicitly represents relationship cues between each pair of AUs for each facial display. These relationship cues are encoded in both node features and multi-dimensional edge features of the graph. The results demonstrate that the proposed node and edge feature learning methods extracted reliable task-specific relationship cues for AU recognition, \emph{i.e.}, both CNN and transformer-based backbones have been largely enhanced, and achieved state-of-the-art results on two widely used datasets. Since our graph-based relationship modelling approach can be easily incorporated with standard CNN/transformer backbones, it can be directly applied to enhance the performance of multi-label tasks or tasks whose data contains multiple objects, by explicitly exploring the task-specific relationship cues among labels or objects.

\appendix
\section{Additional Experimental Results}\label{sec:ex_ar}

\begin{table}[htb]
    \centering
    \small
    \setlength{\tabcolsep}{2.0mm}{
    \begin{tabular}{c|cc|cc}
    \Xhline{3\arrayrulewidth}
     AU  &DRML   &SRERL  &Ours (Res)  & Ours (Swin)  \\
    \midrule
     1 &55.7  &67.6 &75.0 &\textbf{77.7}   \\
     2 &54.5  &70.0 &\textbf{78.0} &76.5   \\
     4 &58.8  &73.4 &85.4 &\textbf{86.5}   \\
     6 &56.6  &78.4 &88.9 &\textbf{89.2}   \\
     7 &61.0  &76.1 &\textbf{84.0} &83.3   \\
     10 &53.6  &80.0 &\textbf{87.4} &86.5   \\
     12 &60.8  &85.9 &93.2 &\textbf{94.0}   \\
     14 &57.0  &64.4 &69.9 &\textbf{73.1}   \\
     15 &56.2  &75.1 &82.7 &\textbf{84.6}   \\
     17 &50.0  &71.7 &\textbf{79.1} &78.7   \\
     23 &53.9  &71.6 &79.5 &\textbf{80.8}   \\
     24 &53.9  &74.6 &\textbf{87.8} &86.3   \\ \midrule
     \textbf{Avg.} &56.0  &74.1 &82.6 &\textbf{83.1}   \\

   \Xhline{3\arrayrulewidth}
    \end{tabular}
    }
    \caption{AUC results achieved for 12 AUs on BP4D dataset.}
    \label{ex:tab_AUC}
\end{table}

\begin{table}[h!]
    \centering
    \small
    \setlength{\tabcolsep}{2.0mm}{
    \begin{tabular}{c|cc|cc}
    \Xhline{3\arrayrulewidth}
     AU  &DRML   &SRERL  &Ours (Res)  & Ours (Swin)  \\
    \midrule
     1 &53.3  &76.2 &\textbf{90.0} &88.6   \\
     2 &53.2  &80.9 &88.5 &\textbf{89.0}   \\
     4 &60.0  &79.1 &\textbf{94.2} &92.3   \\
     6 &54.9  &80.4 &\textbf{92.5} &91.5   \\
     9 &51.5  &76.5 &91.5 &\textbf{92.7}   \\
     12 &54.6  &87.9 &\textbf{95.9} &95.1   \\
     25 &45.6  &90.9 &\textbf{99.1} &98.5   \\
     26 &45.3  &73.4 &\textbf{91.2} &89.4   \\  \midrule
     \textbf{Avg.} &52.3  &80.7 &\textbf{92.9} &92.1   \\

   \Xhline{3\arrayrulewidth}
    \end{tabular}
    }
    \caption{AUC results achieved for 8 AUs on DISFA dataset.}
    \label{ex:tab_AUC}
\end{table}

\section*{Acknowledgements}
Cheng Luo, Weicheng Xie and Linlin Shen are supported by the National Natural Science Foundation of China under grants no. 61602315, 91959108, the Science and Technology Project of Guangdong Province under grant no. 2020A1515010707, the Science and Technology Innovation Commission of Shenzhen under grant no. JCYJ20190808165203670. Siyang Song is supported by the European Union's Horizon~$2020$ Research and Innovation programme under grant agreement No.~$826232$. Hatice Gunes is supported by the EPSRC under grant ref. EP/R$030782$/$1$.

\begin{acronym}
    \acro{AU}{Action Unit}
    \acro{CNN}{Convolutional Neural Network}
    \acro{FACS}{Facial Action Coding System}
    \acro{GCN}{Graph-Convolutional Network}
\end{acronym}

%% The file named.bst is a bibliography style file for BibTeX 0.99c

\bibliographystyle{named}
\bibliography{ijcai22}

\end{document}